\documentclass[a4paper,fleqn]{cas-dc}

\usepackage[authoryear,longnamesfirst]{natbib}
\usepackage{algorithmic}
\usepackage{algorithm}
\usepackage{array}
\usepackage[caption=false,font=normalsize,labelfont=sf,textfont=sf]{subfig}
\usepackage{graphicx}
\usepackage{tabularx}

\usepackage{amsmath,amsfonts,bm}

\def\eqref#1{equation~\ref{#1}}

\def\1{\bm{1}}

\def\va{{\bm{a}}}

\def\vx{{\bm{x}}}
\def\vy{{\bm{y}}}

\def\mA{{\bm{A}}}

\def\mC{{\bm{C}}}

\def\mE{{\bm{E}}}

\def\mH{{\bm{H}}}
\def\mI{{\bm{I}}}
\def\mJ{{\bm{J}}}
\def\mK{{\bm{K}}}
\def\mL{{\bm{L}}}
\def\mM{{\bm{M}}}

\def\mX{{\bm{X}}}

\DeclareMathAlphabet{\mathsfit}{\encodingdefault}{\sfdefault}{m}{sl}
\SetMathAlphabet{\mathsfit}{bold}{\encodingdefault}{\sfdefault}{bx}{n}

\newcommand{\R}{\mathbb{R}}

\newcommand{\KL}{D_{\mathrm{KL}}}

\usepackage{hyperref}
\hypersetup{
    colorlinks=true,
    linkcolor=black,     
    urlcolor=black,
    }
\usepackage{academicons}
\usepackage{xcolor}
\usepackage{color, soul}

\def\tsc#1{\csdef{#1}{\textsc{\lowercase{#1}}\xspace}}
\tsc{WGM}
\tsc{QE}

\begin{document}
\let\WriteBookmarks\relax
\def\floatpagepagefraction{1}
\def\textpagefraction{.001}

\shorttitle{Feature Structure Distillation with Centered Kernel Alignment in BERT Transferring}    

\shortauthors{H. Jung, D. Kim, S. Na, K. Kim}  

\title [mode = title]{Feature Structure Distillation with Centered Kernel Alignment in BERT Transferring}  



%

\author[1]{Hee-Jun Jung}[type=editor,
        orcid=0000-0002-9805-8192]


\fnmark[1]


\ead{jungheejun93@gm.gist.ac.kr}


\credit{<Credit authorship details>}

\affiliation[1]{organization={AI graduate school of Gwangju Institute of Science and Technology (GIST)},
            addressline={123, Cheomdangwagi-ro, Buk-gu,}, 
            city={Gwangju},
            postcode={61005}, 
            country={South Korea}}

\author[1]{Doyeon Kim}[type=editor,
        orcid=0000-0001-7728-3414]



\ead{marri1818@gm.gist.ac.kr}


\credit{<Credit authorship details>}

\author[2]{Seung-Hoon Na}[type=editor,
        orcid=0000-0002-4372-7125]


\ead{nash@jbnu.ac.kr}


\credit{<Credit authorship details>}

\affiliation[2]{organization={Department of Computer Science and Engineering, Jeonbook National University},
            addressline={567, Baekje-daero, Deokjin-gu, Jeonju-si, Jeollabuk-do}, 
            postcode={54896}, 
            country={South Korea}}
            
\author[1]{Kangil Kim}[type=editor,
        orcid=0000-0003-3220-6401]


\cormark[1]

\ead{kikim01@gist.ac.kr}


\credit{}

\cortext[1]{Corresponding author}

\credit{<Credit authorship details>}


\begin{abstract}
Knowledge distillation is an approach to transfer information on representations from a teacher to a student by reducing their difference.
A challenge of this approach is to reduce the flexibility of the student's representations inducing inaccurate learning of the teacher's knowledge.
To resolve it in 
transferring, we investigate distillation of structures of representations specified to three types: intra-feature, local inter-feature, global inter-feature structures.
To transfer them, we introduce \textit{feature structure distillation} methods based on the Centered Kernel Alignment, which assigns a consistent value to similar features structures and reveals more informative relations.
In particular, a memory-augmented transfer method with clustering is implemented for the global structures.
The methods are empirically analyzed on the nine tasks for language understanding of the GLUE dataset with Bidirectional Encoder Representations from Transformers (BERT), which is a representative neural language model.
In the results, the proposed methods effectively transfer the three types of structures and improve performance compared to state-of-the-art distillation methods.
Indeed, the code for the methods is available in \url{https://github.com/maroo-sky/FSD}.
\end{abstract}


\begin{highlights}
\item We adapt CKA to KD for more informative transfer of structures in BERT.
\item We categorize three feature structures (intra-feature, local inter-feature, and global inter-feature structure).
\item We propose their distillation methods, especially memory augmentation with clustering for global structures.
\item We empirically analyze restoration rate, patterns of transferring feature structures, and task-specific properties.
\item We validate practical usefulness over a wide range of language understanding tasks (GLUE benchmark).  
\end{highlights}


\maketitle

\section{Introduction}\label{}
{I}{n} current deep learning models, knowledge distillation (KD) is a common approach to transfer information of features of a larger model to a smaller student model~\cite{survey}. 
This approach reduces the difference in prediction confidence between two models.
Confidence is usually represented as a probability vector. 
Moreover, distillation can be applied to various vector distributions to transfer more feature information. 
For example, distribution on an intermediate layer~\cite{sun-etal-2019-patient, wang2020minilm} or a final fully connected layer~\cite{DBLP:journals/corr/abs-1909-10351} have been directly compared. To transfer more rich information, pairwise relations between features~\cite{9009071, DBLP:journals/corr/abs-1904-05068, 10.1007/978-3-030-58610-2_2} have also been used. 
A problem with the direct fitting of a vector is its huge flexibility in geometric space, even in the same setting of neural networks.
This flexibility may cause ambiguity to the guide for a student to learn the teacher's knowledge. 

Transferring more rich information on representations' connectivity is a possible solution. 
Centered Kernel Alignment (CKA)~\cite{DBLP:journals/corr/abs-1203-0550} is a suitable metric for this approach as it assigns a similarity value to feature structure. 
Furthermore, its score is more consistent on potentially similar representations trained on different architectures and layers~\cite{DBLP:journals/corr/abs-1905-00414}.
This property is expected to help distillation focus on more informative feature distribution.
Implementations of this approach have been reported in a few recent computer vision tasks, but widely used BERT model in natural language processing is not sufficiently studied yet. 


In this work, we propose \textit{feature structure distillation} (FSD) method to adapt CKA to KD between a teacher and a student model. 
The proposed methods transfer rich information categorized into three types of structures on the feature representations: intra-feature, local inter-feature, and global inter-feature structures.
A separate distillation loss is introduced for each structure defined on the feature distribution generated from the penultimate layer. 
To obtain the global inter-feature structures over the full batch of training samples, we newly add a memory architecture that is induced via clustering. 

We present experiments on the General Language Understanding Evaluation (GLUE)~\cite{wang2018glue} benchmark with BERT distilled by FSD methods. 
In the results, FSD methods show possibility that these methods outperform other state-of-the-art KD methods and even teacher models in some tasks.
Far from many previous works \cite{DBLP:journals/corr/abs-1909-10351, sun-etal-2020-mobilebert, wang2020minilm, sun-etal-2019-patient, 9009071, DBLP:journals/corr/abs-1904-05068, 10.1007/978-3-030-58610-2_2, context-distl} which mainly focus on model performance, we provide the results of the restoration rate of the teacher's prediction and the similarity change of geometric structures for deeper understanding of the structure distillation. 

Our key contributions are summarized as follows: 
\begin{itemize}
    \item We adapt CKA to KD for more informative transfer of structures in BERT.
    \item We categorize three feature structures (intra-feature, local inter-feature, and global inter-feature structure).
    \item We propose their distillation methods, especially memory augmentation with clustering for global structures.
    \item We empirically analyze restoration rate, patterns of transferring feature structures, and task-specific properties.
    \item We validate practical usefulness over a wide range of language understanding tasks (GLUE benchmark).  
\end{itemize}

\begin{table}[h!]
    \footnotesize
    \centering
    \begin{tabular}{p{0.12\textwidth} p{0.3\textwidth}}
    \multicolumn{2}{c}{Terms and Notations}\\
    \hline
    feature & a representation vector \\
    relation & a pair-wise relation of two features \\
    feature structure & a set of relations \\
    feature distribution & a set of features\\
    $\displaystyle a$ & A scalar (integer or real)\\
    $\displaystyle \va$ & a vector\\
    $\displaystyle \mA$ & a matrix\\
    $\displaystyle \mI_n$ & identity matrix with $n$ rows and $n$ columns\\
    $\displaystyle \mJ_n$ & all-ones matrix with $n$ rows and $n$ columns\\
    $\displaystyle \R$ & the set of real numbers \\
    $\displaystyle \R^{m \times n}$ & $m$ by $n$ shape of matrix \\
    $\displaystyle \R^{m \times n \times k}$ & $m$ by $n$ by $k$ shape of 3rd-order tensor \\
    $\displaystyle \{0, 1, \dots, n \}$ & the set of all integers between $0$ and $n$\\
    $\displaystyle \KL ( P \Vert Q ) $ & Kullback-Leibler divergence of P and Q \\
    $\displaystyle || \vx ||_2 $ & L2 norm of $\vx$ \\
    FSD & proposed distillation for all feature structure types\\
    FSD$_I$ & FSD for only local intra-feature structure \\
    FSD$_L$ & FSD for only local inter-feature structure \\
    FSD$_G$ & FSD for only global inter-feature structure \\
    FSD$_{IL}$ & integration of FSD$_I$ and FSD$_L$ \\
    \hline
    \label{term_notation}
    \end{tabular}
\end{table}

\section{Related Work}

\subsection{Analysis of Similarity of Representation}
\label{sub:analysis of similarity}
The similarity between representations of deep networks has been measured by various methods. 
Canonical Correlation Analysis (CCA)~\cite{Hotelling1992} estimates the association between two variables and identifies a linear relationship with weight to maximize correlation.
CCA is sensitive to perturbation when the condition number of representations is large~\cite{10.1007/978-1-4612-4228-4_3}.
To reduce the sensitivity of perturbation, Singular Value Canonical correlation Analysis (SVCCA)~\cite{NIPS2017_dc6a7e65} applies singular value decomposition to use more important principal components, and Projection Weighted CCA (PWCCA)~\cite{NEURIPS2018_a7a3d70c} assigns higher weights to more important canonical correlations.
These methods aimed to assign the same relation of flexibly located representations in different models, but the consistency  of their methods is insufficient~\cite{DBLP:journals/corr/abs-1905-00414}.
CKA is an alternative for enhancing the invariance to orthogonal transformation and isotropic scaling, which is expected to enhance the consistency~\cite{DBLP:journals/corr/abs-1905-00414}
The metric improved the performance of alignment-based algorithms~\cite{DBLP:journals/corr/abs-1203-0550}, measuring the similarity between kernels or kernel matrices. 
Furthermore, CKA outperforms CCA, SVCCA, and PWCCA on the test of identifying corresponding layers~\cite{DBLP:journals/corr/abs-1905-00414}. 




\subsection{Knowledge Distillation for BERT}
\label{sub:KD}
KD~\cite{hinton2015distilling} is a method to transfer dark knowledge of a large teacher model to a smaller student model while preserving the training accuracy, and this method is applied to DistilBERT~\cite{DBLP:journals/corr/abs-1910-01108}.
An extension of KD is to directly reduce distance between representations.
For example, 
TinyBERT~\cite{DBLP:journals/corr/abs-1909-10351} delivers word embedding, self-attention head, and representations on selected intermediate layers.
MobileBERT~\cite{sun-etal-2020-mobilebert} moves representations on all layers to a student of the same number of layers, and MiniLM~\cite{wang2020minilm} uses relations between the values in the self-attention and the attention distribution that is computed from the scaled dot products of the queries and keys~\cite{NIPS2017_3f5ee243}.
DistilBERT and TinyBERT models perform distillation on the pre-training and fine-tuning stages but MobileBERT and MiniLM models operate distillation only on the pre-training stage.
To reduce the interference of other factors in the analysis, we conduct distillation on the fine-tuning stage as~\textit{patient knowledge distillation}  (PKD)~\cite{sun-etal-2019-patient}. 
The method implements teacher representations from multiple intermediate layers normalized to the student layers in downstream tasks, enabling transfer between neural networks of different numbers of layers.
These methods penalize the difference between teacher and student features, then force them closer in the same vector space.
However, CKA allows of using different vector spaces and dimensions.

\subsection{Transferring Rich Information}
Transferring rich information as differences or relation has been introduced in a few vision tasks. 
Correlation Congruence for Knowledge Distillation (CCKD)~\cite{9009071} transfers a correlation matrix between representations generated from kernels, Relational Knowledge Distillation (RKD)~\cite{DBLP:journals/corr/abs-1904-05068} evaluates the difference in Euclidean distance or cosine similarity, and Local Correlation Consistency for Knowledge Distillation (LKD)~\cite{10.1007/978-3-030-58610-2_2} additionally uses the difference of angles, each of which is determined by three representations.
Indeed, Contextual Knowledge Distillation CKD \cite{context-distl} proposed layer transforming relation as well as word relation-based contextual knowledge distillation with same manner of RKD to evaluate the difference of structure.
It extends transferring teacher structure from word level (within same layer) to layer level (over layers).
In addition, Similarity-Preserving Knowledge Distillation~\cite{Tung_2019_ICCV} transfers pairwise similarity with outer products of mini-batch samples.
Another approach~\cite{IRG} conveys teacher knowledge with Instance Relationship Graph (IRG) to student for reducing the distance of vertex by vertex and edge by edge of IRG.
Differently, our methods cover a wider range of feature structures as global structures and effective intra-feature structures are specifically designed for transformers. 
To measure the difference between structures, we adopt CKA, which effectively maintains the implicit relations between representations~\cite{DBLP:journals/corr/abs-1905-00414}.
CKA has been adopted for KD in convolutional networks~\cite{10.1007/978-3-030-59710-8_71} and has shown successful performance improvement, but the extension to global structure and a deeper analysis in the transformer networks have rarely been discussed. 


\begin{figure*}[ht]
    \centering
    \begin{subfloat}[]{\includegraphics[width=0.42\textwidth]{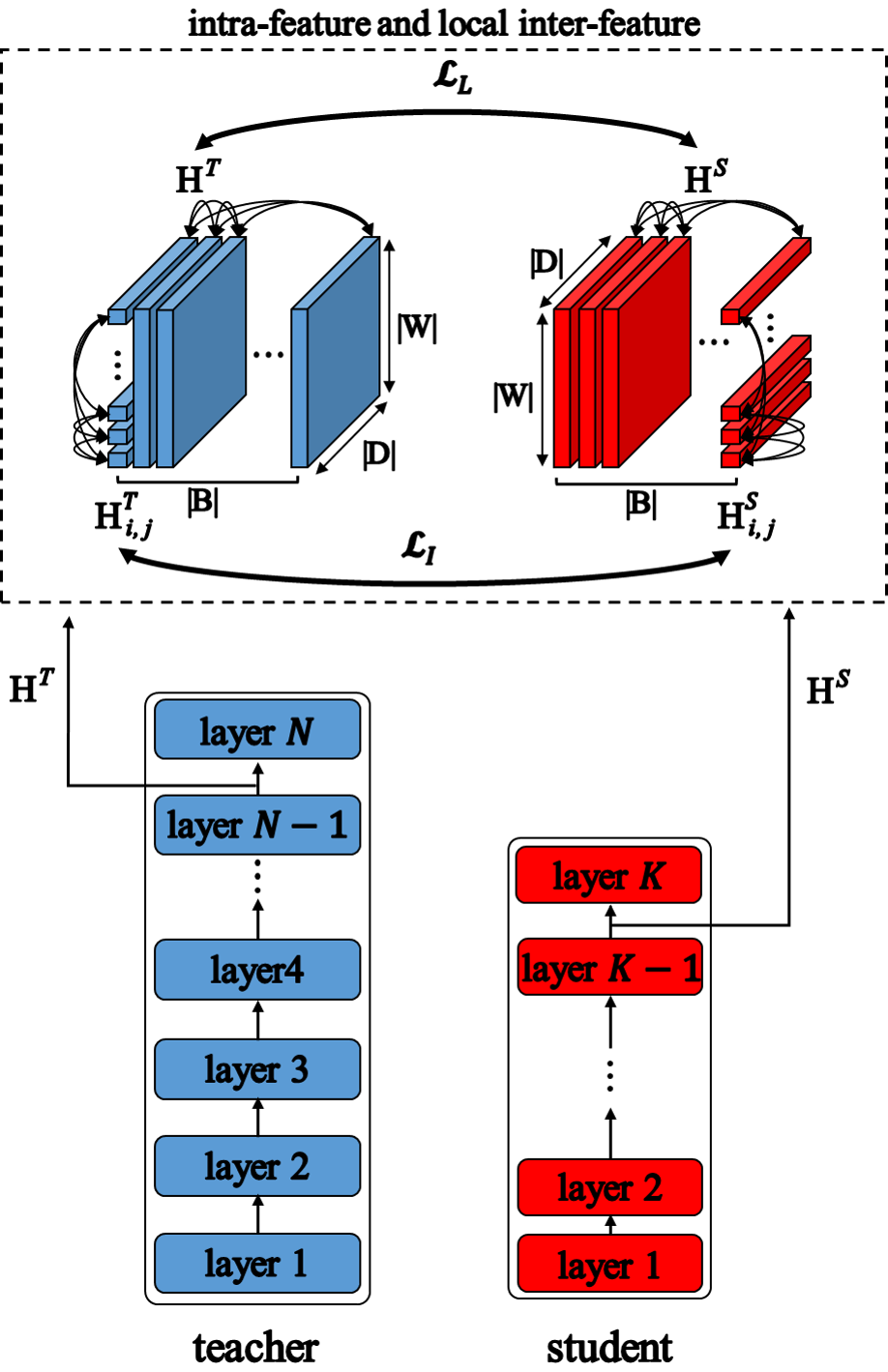}
        \label{subfig:inter-inter structure tansfer}}
    \end{subfloat}
    \hfill
    \begin{subfloat}[]{\includegraphics[width=0.55\textwidth]{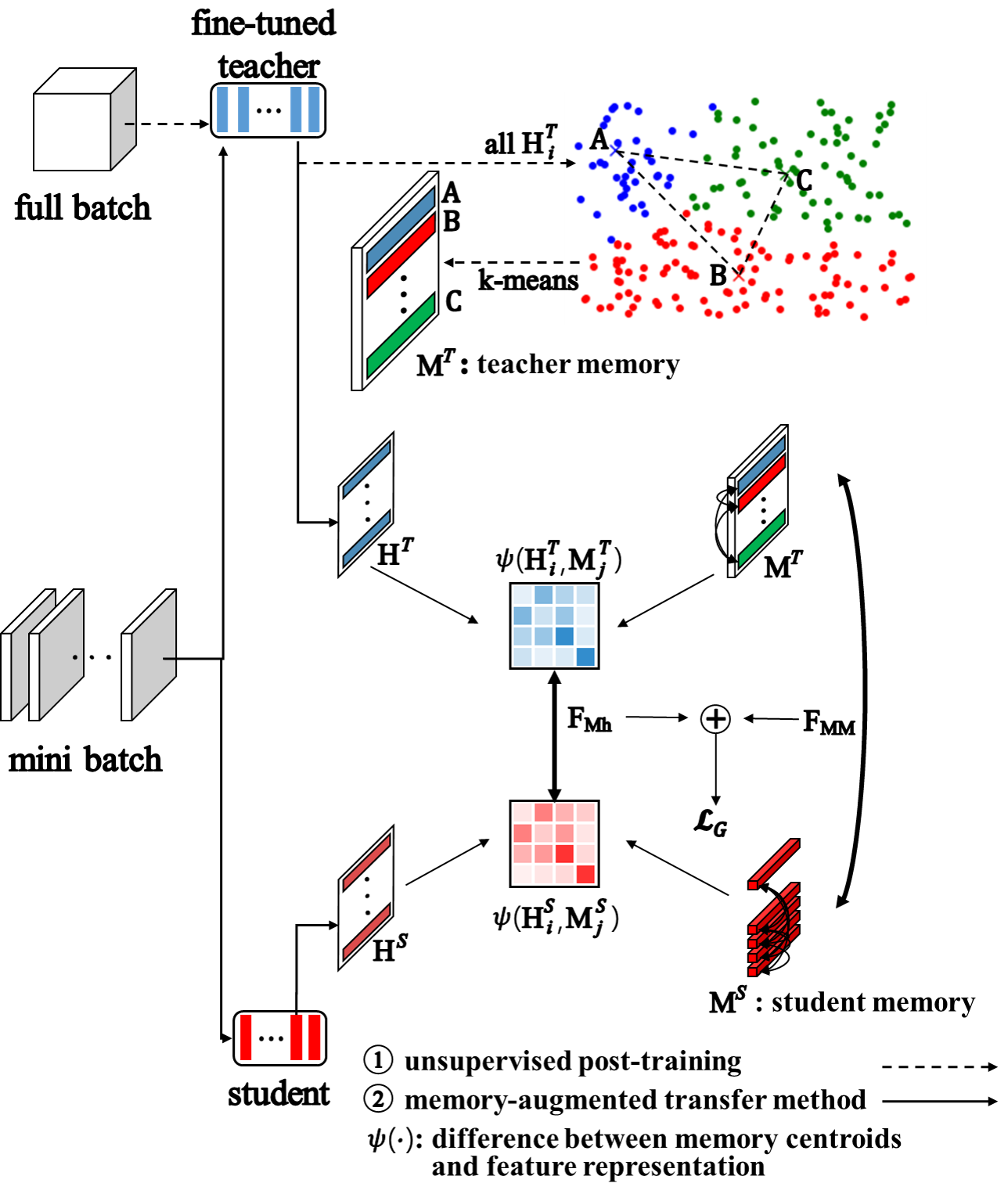}
    \label{subfig:global-inter structure transfer}}
    
    \end{subfloat}
    \caption{Overview of the proposed loss functions of distillation methods for three feature structures: $\mathcal{L}_{I}$ and $\mathcal{L}_{L}$ in Figure~\ref{subfig:inter-inter structure tansfer} uses CKA between teacher and student representations on the penultimate layer. 
    In Figure~\ref{subfig:global-inter structure transfer}, the first stage is an unsupervised post-training and the second stage is a memory-augmented transfer method.
    In the first stage, $\displaystyle \mM^T$ is first trained to memorize the centroids of representations on the penultimate layer of a teacher model. Then, its stored centroids are transferred to a student model by $\mathcal{L}_{G}$ in the second stage.}
    \label{fig:overview of the proposed methods}
\end{figure*}

\section{Method}
\label{sec:method}
In this section, we clarify the KD settings for presenting the FSD methods. 

\subsection{Base Knowledge Distillation}
We set two compatible base settings for KD to transfer a feature distribution introduced in previous works~\cite{hinton2015distilling,sun-etal-2019-patient}.
In the former setting, given $N$ training samples for training a student model $S$, a fine-tuned teacher model $T$ transfers a feature distribution on the final layer by training $S$ with the similarity loss as
\begin{equation}
    \label{eq:kldivergence}
    \mathcal{L}_{KLD} = \sum_{i=1}^N \displaystyle \KL ( \frac{\emph{f}_T(\displaystyle \vx_i)}{\displaystyle \tau} \Vert \frac{\emph{f}_S(\displaystyle \vx_i)}{\displaystyle \tau}),
\end{equation}
where 
$\emph{f}(\cdot)$ is a neural network to generate a probability vector through softmax function and layers.
An input sample for the network is composed of vector $\displaystyle \vx_i$ and its ground-truth $\displaystyle \vy_i$, where $\displaystyle i$ is the sample index in the training data. 
The temperature $\displaystyle \tau$ is to control relaxation. 
To distill teacher knowledge, the model is trained by a task-specific cross-entropy loss $\mathcal{L}_{CE}$ with  $\mathcal{L}_{KLD}$ as 
\begin{equation}
    \label{eq: vkd objective function}
    \mathcal{L}_{VKD} = \displaystyle \alpha\mathcal{L}_{CE} + (1 - \displaystyle \alpha)\mathcal{L}_{KLD},
\end{equation}
where $\alpha$ is the interpolation rate between two loss functions, which needs to be tuned empirically.
This approach to transfer a feature with the label is called \textit{vanilla knowledge distillation} (VKD) in this paper.
The latter setting follows the Eq.~\ref{eq: vkd objective function}, but the similarity loss is replaced with the distance between hidden vectors generated from the several intermediate layers and penultimate layers $f^\prime(\cdot)$ of $T$ and $S$, which is the method of PKD~\cite{sun-etal-2019-patient}.
As the setting of both methods, we initialize the parameters of $S$ from the pre-trained $T$ and then perform distillation on the fine-tuning stage.

In this paper, the proposed FSD method is used as distillation loss functions in training the student model as 
\begin{equation}
    \label{eq:intra-feature objective function}
    \mathcal{L} = \mathcal{L}_{VKD} + \displaystyle \beta\mathcal{L}_{FSD},
\end{equation}
where $\displaystyle \beta$ is a hyper-parameter to control the interpolation rate between VKD loss and the proposed distillation loss. 

\subsection{Feature Structure Distillation with CKA}
\label{sub:structure transfer}
\subsubsection{Overview}
In this paper, we propose FSD method to transfer more rich information on representations by comparing feature structures relations rather than relation. 
\textit{Feature structure} is split into three groups with respect to their locality: intra-feature, local inter-feature, and global inter-feature structures.
Fig.~\ref{fig:overview of the proposed methods} shows the distinction of the structures.
In addition, feature structures are conveyed only on the penultimate layer to compare baselines (PKD and RKD).


\subsubsection{Similarity Between Feature Structures}
We adapt CKA for evaluating similarity between feature structures in order to use its robustness to the flexibility of feature distribution and consequently to reduce ambiguity of distillation.
\begin{equation}
    \label{eq:CKA general form}
    \emph{CKA}(\displaystyle \mE^1,\displaystyle \mE^2) = \frac{\emph{HSIC}(\displaystyle \mK, \displaystyle \mL)}{\sqrt{\emph{HSIC}(\displaystyle \mK, \displaystyle \mK)\emph{HSIC}(\displaystyle \mL, \displaystyle \mL)}},
\end{equation}
where $\displaystyle \mE^1$ and $\displaystyle \mE^2$ are features; the function $\emph{HSIC}$ is the Hilbert-Schmidt Independence Criterion for determining Independence of two sets of variables ~\cite{10.1007/11564089_7}; $\displaystyle \mK = {\displaystyle \mE^1}^\textsc{t} \displaystyle \mE^1$, $\displaystyle \mL = {\displaystyle \mE^2}^\textsc{t} \displaystyle \mE^2$.
The function HSIC is the Hilbert-Schmidt Independence Criterion~\cite{10.1007/11564089_7}  defined as
\begin{equation}
    \label{eq:HSIC}
    \emph{HSIC}(\displaystyle \mK, \displaystyle \mL) = \frac{1}{(\displaystyle N -1)^2}tr(\displaystyle \mK \displaystyle \mC \displaystyle \mL \displaystyle \mC),
\end{equation}
where $tr$ is a trace in a matrix, $\displaystyle \mC$ is a centering matrix $\displaystyle \mC_{\displaystyle n} = 
\displaystyle \mI_{\displaystyle n} - \frac{1}{\displaystyle n} \displaystyle \mJ_{\displaystyle n}$.

In each proposed method, we use different $\displaystyle \mE^1$ and $\displaystyle \mE^2$, but they are all based on the hidden vectors generated from the penultimate layer of teacher, and student, notated as $\displaystyle \mH^T$ for the teacher and $\displaystyle \mH^S$ for the student in the shape of $\mathbb{R}^{|B|\times|W||D|}$. The constants $|\displaystyle B|$, $|\displaystyle W|$, and $|\displaystyle D|$ are the number of samples in a mini-batch, the maximum sequence length, and the hidden state dimension, respectively.

\subsubsection{Intra-Feature Structure Distillation (FSD$_I$)}
\textit{intra-feature structure} implies the set of difference values between segments of the hidden vector from the penultimate layer from a single input sample. 
In the transformer networks, the unit for segmentation is a token. 
To obtain the difference of token-level structures between the teacher and student model, we split the hidden vector $\displaystyle \mH^S_{i} \in \displaystyle \R^{|W||D|}$ into token-level feature vectors 
$\displaystyle \mH^S_{i,j} \in \displaystyle \R^{|D|}$ for the $j$th token from $i$th input sample and this is equally applied to the teacher model.
Then, we reshape teacher and student penultimate layer representation $\displaystyle \mH^T$ and $\displaystyle \mH^S$ into $\displaystyle \mH^T_{re}$ and $\displaystyle \mH^S_{re}$ in shape of $\mathbb{R}^{|B| \times |W| \times |D|}$.
The loss function implying the difference is defined as
\begin{equation}
\label{eq:Intra loss}
    \mathcal{L}_I = -\frac{1}{|B|}\sum_{i=1}^{|B|}\log \emph{CKA}|(\displaystyle \mH^S_{re},\displaystyle \mH^T_{re})|.
\end{equation}

\subsubsection{Local Inter-Feature Structure Distillation (FSD$_L$)}
\textit{Local inter-feature structure} implies the set of difference between hidden vectors generated at the penultimate layer from the samples of a mini-batch. 
Compared to the intra-feature structure method, it deals with the structure between samples rather than internal units from a single sample. 
The distillation loss adapting CKA for comparing the local inter-feature structures is 
\begin{eqnarray}
\label{eq:Inter loss}
    \mathcal{L}_L &=& -\log{|\emph{CKA}(\displaystyle \mH^S,\displaystyle \mH^T)|}.
\end{eqnarray}

\subsubsection{Global Inter-Feature Structure Distillation via Memory Augmentation (FSD$_G$)}
\label{sec:global structure}

\begin{figure}[htbp]
    \centering
    \includegraphics[width=0.48\textwidth]{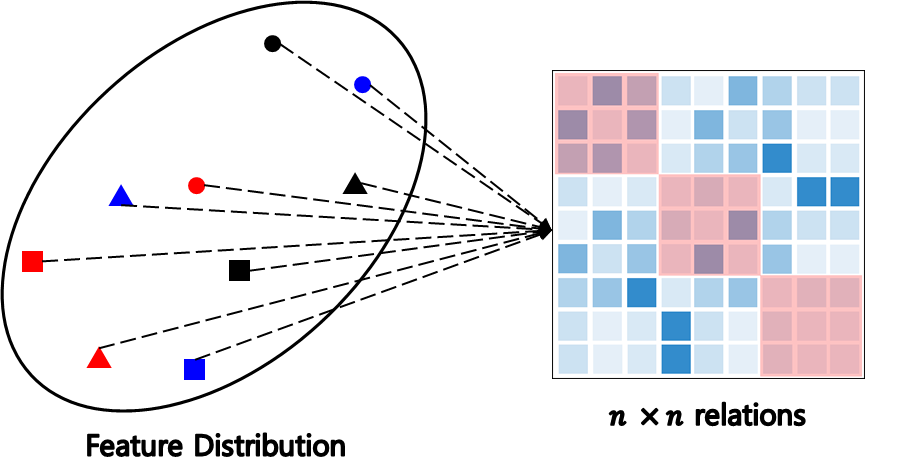}
    \caption{The difference of covered relations between a local structure and a global structure is shown. The blue blocks of the right-side matrix are relations of the global structure in the full batch. The red blocks are found in mini-batches (local structure).
    Variously colored and shaped markers indicate features and the matrix shows relations over the features.}
    \label{fig:memory_size}
\end{figure}

The local approach is easy to implement but transfers only the partial inter-feature structures because it evaluates the structures only for samples in the same mini-batch. 
To transfer all inter-feature structures, a pairwise difference for two samples has to be evaluated for $O(n^2)$ relations, but the local method considers $O(\frac{n}{b}b^2) = O(n)$ where $n$ is the total number of samples in the full batch, and $b$ is a given constant number of samples in a mini-batch.
As shown in Fig.~\ref{fig:memory_size}, relations in mini-batches could not cover all relations in full batches.
Thus, the coverage of FSD$_{L}$ rapidly decreases by the input samples size. 
This decrease implies that the transferred feature structure is easily insufficient for a large dataset.
Moreover, the large scale makes transferring all structures computationally inefficient and infeasible in parallel programming with limited hardware. 

To overcome this problem, we propose an approach to effectively transfer the inter-feature structures between samples in different mini-batches, called \textit{global inter-feature} structures.  
The key idea is to store centroids of full batch into a memory and then transfer the structures between centroids in a teacher and a student. 
As locating features of the student to its centroids, the method transfers the global inter-feature structures built by samples near the involved centroids.
This method has two sequential stages: 1)~\textit{unsupervised post-training} and 2)~\textit{memory-augmented transfer method}.
In the first stage, we use the fine-tuned teacher to generate representations of all samples from the penultimate layer in the full batch.
After randomly initialized element vectors in memory, each vector is updated to minimize the Euclidean distance to its corresponding k-nearest neighbors, thereby performing \textit{k-means clustering}. 
The second stage is the memory-augmented transfer method to reduce the difference of inter-feature structures over the centroids in the teacher memory $\displaystyle \mM^T$ and student memory $\displaystyle \mM^S \in \displaystyle \R
^{(|C|\times|W||D|)}$, where the number of elements of the memory $|\displaystyle C| \in \{100,~300\}$. 
The difference $\textbf{F}_{\mathbf{MM}}$ is defined as following equation:
\begin{equation}
\begin{split}
\label{eq: memory loss}
    \textbf{F}_{\mathbf{MM}} = - \log|\emph{CKA}(\displaystyle \mM^T , \displaystyle \mM^S)|.
\end{split}
\end{equation}
The transferred structures on the centroids are propagated to features of the student by learning the teacher's feature-to-centroid difference.
For transferring centroid difference, we define $\textbf{F}_\mathbf{Mh}$ as 
\begin{equation}
    \label{eq:memory-hidden structure loss}
    \textbf{F}_{\mathbf{Mh}} =  \frac{\sum_{i,j} ||\psi(\displaystyle \mH^T_{i}, \displaystyle \mM^T_{j}) - \psi(\displaystyle \mH^S_{i}, \displaystyle \mM^S_{j})||_2^2}{|B||C|},
\end{equation}
where $\psi$ is a function for evaluating distance. $\textbf{F}^E_{\mathbf{Mh}}$ and $\textbf{F}^C_{\mathbf{Mh}}$ specify $\psi$ to Euclidean distance and cosine similarity, respectively.
The final distillation loss for transferring the global inter-feature structure is defined as follows:
\begin{equation}
    \label{eq:global structure loss}
    \mathcal{L}_{G}= \gamma_m \textbf{F}_{\mathbf{Mh}}^E +  (1 - \gamma_m) \textbf{F}_{\mathbf{Mh}}^C + \textbf{F}_{\mathbf{MM}},
\end{equation}
where $\gamma_m$ is a hyper-parameter for interpolation.

We used Euclidean and cosine similarity as ~\cite{DBLP:journals/corr/abs-1904-05068, context-distl} instead of CKA for $\textbf{F}_\mathbf{Mh}$ to allow this method to set a flexible memory size because CKA can not be directly applied when the batch size of $\mH$ is different to the number of centroids in the memory.


\subsubsection{Integration}
The intra-feature, local inter-feature, and global inter-feature structures represent different types of structures. 
Thus, their integration is a natural extension for transferring more rich feature structure.
The integration is simply implemented as the sum of all three loss functions as:
\begin{equation}
    \mathcal{L}_{FSD} = 
    \gamma_i \mathcal{L}_{I} + \gamma_l \mathcal{L}_{L} + \gamma_g \mathcal{L}_{G},
\end{equation}
where the hyper-parameter $\gamma_i$, $\gamma_l$, and $\gamma_g$ are interpolation rates.
This integrated loss is used for training a student model as described in the following Algorithm~\ref{alg:alg1}.
\begin{algorithm}[H]
\caption{FSD Loss}\label{alg:alg1}
\begin{algorithmic}
\STATE
\REQUIRE $i$-th iteration mini-batch $\displaystyle \mX_i \in \mathcal{X}$ (dataset) $\displaystyle \mX_i \in~\mathbb{R}^{|B| \times |W| \times |D|}$, hyper-parameters $\beta, \gamma_m, \gamma_i, \gamma_l$, $\gamma_g$, and learning rate $\eta$. $M$ is a memory, $f^H(\cdot)$ is output of pernultimate layer. $S$ and $T$ is student and teacher respectively.
\ENSURE Loss $\mathcal{L}_{FSD}$
\STATE
\STATE $\texttt{\#}$ We freeze the teacher memory $M^T$ during the training. 
\STATE
\STATE reshape($\cdot$): $\mathbb{R}^{|B| \times |W| \times |D|} \rightarrow \mathbb{R}^{|B| \times |W||D|}$
\STATE $ \displaystyle \mH_{re}^S, \displaystyle \mH_{re}^T \leftarrow f_S^H (\displaystyle \mX_i), f_T^H (\displaystyle \mX_i)$
\STATE  $\mathcal{L}_I \leftarrow -\frac{1}{|B|}\sum_{i=1}^{|B|}\log \emph{CKA}|(\displaystyle \mH^S_{re},\displaystyle \mH^T_{re})|$ ~~~~~~~~~~\COMMENT{Eq.~\ref{eq:Intra loss}}
\STATE $ \displaystyle \mH^S, \displaystyle \mH^T \leftarrow \text{reshape}(\displaystyle \mH_{re}^S), ~\text{reshape}(\displaystyle \mH_{re}^T)$
\STATE  $\mathcal{L}_L \leftarrow -\log{|\emph{CKA}(\displaystyle \mH^S,\displaystyle \mH^T)|}$~~~~~~~~~~~~~~~~~~~~~~~~\COMMENT{Eq.~\ref{eq:Inter loss}}
\STATE  $\textbf{F}_{\mathbf{MM}} \leftarrow - \log|\emph{CKA}(\displaystyle \mM^T , \displaystyle \mM^S)|$ ~~~~~~~~~~~~~~~~~~\COMMENT{Eq.~\ref{eq: memory loss}}
\STATE  $\textbf{F}_{\mathbf{Mh}} \leftarrow  \frac{\sum_{i,j} ||\psi(\displaystyle \mH^T_{i}, \displaystyle \mM^T_{j}) - \psi(\displaystyle \mH^S_{i}, \displaystyle \mM^S_{j})||_2^2}{|B||C|}$  ~~~\COMMENT{Eq.~\ref{eq:memory-hidden structure loss}}
\STATE  $\mathcal{L}_{G} \leftarrow \gamma_m \textbf{F}_{\mathbf{Mh}}^E +  (1 - \gamma_m) \textbf{F}_{\mathbf{Mh}}^C + \textbf{F}_{\mathbf{MM}}$  ~~~~~~~~~~\COMMENT{Eq.~\ref{eq:global structure loss}}
\STATE  $\mathcal{L}_{FSD} \leftarrow \gamma_i \mathcal{L}_{I} + \gamma_l \mathcal{L}_{L} + \gamma_g\mathcal{L}_{G}$
\RETURN $\mathcal{L}_{FSD}$
\end{algorithmic}
\label{alg1}
\end{algorithm}

\section{Experiments}
We empirically analyze FSD method applied to BERT for well-known language understanding tasks.
Beyond usual performance and computational efficiency evaluation to evaluate practical impact, we focus on understanding of how teacher knowledge is effectively transferred, because its impact to performance is promising as shown in~\cite{xu-etal-2021-beyond} and the final performance is interfered by other effects of distillation as generalization~\cite{Yuan2020RevisitingKD, heeseung}.
The analysis has three parts 1) quantitative analysis, 2) qualitative analysis, 3) and additional discussion.
In the quantitative analysis, we evaluate the practical impact of our method compared with state-of-the-art, the impact of each feature structure level, and the impact of memory stability.
In the qualitative analysis, we analyze how much student reflects and is close to teacher's structure.
In the last, we discuss geometry property, and model and time complexity.
The performance is for evaluating practical impact of our methods in comparison with state-of-the-art.
The other three parts are to evaluate the effectiveness of transferring teacher knowledge on feature distributions in various perspectives.

\subsection{Datasets}
\label{sub:datasets}
The General Language Understanding Evaluation (GLUE)~\footnotemark{} benchmark is presented in Table~\ref{glue}
\begin{table}[h!]
    \scriptsize
    \centering
    \caption{Evaluation Metrics and the number of dataset of GLUE benchmark. |\textbf{train}| and |\textbf{dev}| are the size of training and development dataset, and Corr is a correlation.}
    \begin{tabular}{llccc}
    \hline
     &\textbf{corpus} & \textbf{$|$train$|$} & \textbf{$|$dev$|$} & \textbf{metrics} \\ 
     \hline
     \multirow{2}{*}{\shortstack{single-sentence \\tasks}} &
     CoLA & 8.5k & 1k & Matthews Corr \\
     & SST-2 & 67k & 872 & Accuracy \\
     \hline
     \multirow{4}{*}{\shortstack{similarity and \\paraphrase tasks}} &
     QQP & 364k & 40k & Accuracy/F1 \\
     & MRPC & 3.7k & 408 & Accuracy/F1 \\
     & \multirow{2}{*}{STS-B} & \multirow{2}{*}{7k} & \multirow{2}{*}{1.5k} & Pearson Corr \\
     & \multirow{2}{*}{} & \multirow{2}{*}{} & \multirow{2}{*}{} & Spearman Corr \\
     \hline
     \multirow{4}{*}{inference tasks} &
     MNLI & 393k & 20k & Accuracy\\
     & RTE & 2.5k & 276 & Accuracy\\
     & QNLI & 105k & 5.5k & Accuracy\\
     & WNLI & 634 & 71 & Accuracy\\
     \hline
    \end{tabular}

    \label{glue}
\end{table}
\footnotetext{https://gluebenchmark.com/tasks}

\textbf{GLUE} the General Language Understanding Evaluation~\cite{wang2018glue} consists of nine English sentence-understanding tasks.
Single-sentence tasks include the Corpus of Linguistic Acceptability (CoLA)~\cite{DBLP:journals/corr/abs-1805-12471}, Stanford Sentiment Treebank~\cite{socher-etal-2013-recursive}.
In the similarity and paraphrase tasks, Microsoft Research Paraphrase Corpus (MRPC)~\cite{dolan-brockett-2005-automatically}, Quora Question Pairs (QQP)\footnote{https://data.quora.com/First-Quora-Dataset-Release-
Question-Pairs}, and Semantic Textual Similarity Benchmark (STS-B)~\cite{cer-etal-2017-semeval} are included.
In the last, inference tasks include the Multi-Genre Natural Language Inference Corpus (MNLI)~\cite{williams-etal-2018-broad}, Stanford Question Answering Dataset (QNLI) ~\cite{rajpurkar-etal-2016-squad}, Recognizing Textual Entailment (RTE) ~\cite{10.1007/11736790_9, 10.5555/1654536.1654538, Bentivogli09thefifth}, and Winograd Schema Challenge (WNLI)~\cite{lev11}.


\subsection{Distillation Settings}
\subsubsection{Environment Setup}
We conduct knowledge distillation with GLUE on a single RTX-2080-Ti and RTX-8000 GPU with 32 batches, 
128 max sequence length, and 768 dimensions.
WNLI, MPRC, and SST-2 are implemented on single RTX-2080-Ti, and RTE, STS-B, CoLA, and QNLI are implemented on a single RTX-8000 GPU. 
FSD of QQP and MNLI are implemented on a single RTX-8000 because of the memory size, and other methods of QQP and MNLI are operated on a single RTX-2080-Ti GPU.
Each task performance is slightly different depend on GPU device and number of device.

\subsubsection{BERT-base Preparation}
We set a 12-layer transformer encoder with 768 hidden nodes and 12 attention heads as a teacher model.
We conduct fine-tuning with the uncased version of pre-trained BERT-base\footnote{https://s3.amazonaws.com/models.huggingface.co/
bert/bertbase-uncased-pytorch model.bin} on nine GLUE tasks independently.
The maximum sequence length is 128 which is referred in~\cite{sun-etal-2019-patient}.
The number of train epochs is 3.
The training batch size is 32.
The learning rate is 2e-5 except for STS-B and WNLI tasks, which are set in 5e-5 to slightly improve teacher performance.
We note that fine-tuning of BERT-base can be more improved by adding other methods irrelevant to knowledge distillation. In this paper, our primary goal is not to solve language understanding tasks by whatever means possible, but to prove the impact of more accurate transferring of teacher’s knowledge in a practical environment. Therefore, fair comparative group is the state-of-the-art transferring methods rather than the state-of-the-art language understanding model.

\subsubsection{Baseline Method Settings}
We reproduce VKD, PKD, MiniLM, and RKD but MiniLM~\cite{wang2020minilm} performs KD on the pre-training stage.
For consistency with VKD, PKD, and FSD, we apply MiniLM method on the fine-tuning stage.
Previous work~\cite{sun-etal-2019-patient} uses 6-layers of BERT model (BERT$_{6}$) as a student and we implement distillation experiments with the same student architecture.
We utilize parameters from 1$^{st}$ to 6$^{th}$ layer of pre-trained BERT$_{BASE}$ to initialize BERT$_6$.
Fine-tuning for VKD, we conduct each task with $\alpha$ from $\{0.2, 0.5, 0.7\}$, temperature $\tau$ from $\{5, 10, 20\}$, and learning rate from $\{$1e-5, 2e-5, 5e-5$\}$ to search for the best model.
Additionally, we set angle and distance loss hyper-parameters introduced in 
To reproduce RKD~\cite{DBLP:journals/corr/abs-1904-05068}, we set hyper-parameters for its angle and distance loss from $\{200.0, 2000.0, 20000.0\}$, and $\{100.0, 1000.0, 10000.0\}$. 

\subsection{FSD Settings}

\begin{table*}[!ht]
\scriptsize
{
    \centering
    \caption{
    Performance on small tasks of GLUE benchmark.
    Mean and standard deviation of their evaluation metrics on six runs on their development (dev.)sets.
    [\textit{p}-value]s of t-test are calculated for comparison of the proposed and the best baseline methods.
    The numbers imply accuracy in WNLI and RTE, Pearson/Spearman correlation in STS-B, Matthew's correlation in CoLA, and F1/accuracy in MRPC task. A bold number is the best result among distillation methods in each task.
    $^\ast$ denotes that the results are taken from the huggingface BERT$_{BERT}$\protect\footnotemark{}. 
    $\dag$ denotes reproduced model for consistency with VKD, PKD, and FSD methods.
    }
    \begin{tabular}{clccccc}
    \hline
    & \textbf{method} & \textbf{WNLI} & \textbf{RTE} & \textbf{STS-B} & \textbf{CoLA} & \textbf{MRPC} \\
    \hline
    \multirow{2}{*}{\textbf{teacher}} & \textbf{HF$^{\ast}$} & 56.34 & 67.15 & 93.95/83.70 & 49.23 & 89.47/85.29\\
    \multirow{2}{*}{} & \textbf{BERT-base} & 56.34 & 66.79 & 88.78/88.48 & 55.47 & 86.00/81.10\\
    \hline
    \multirow{4}{*}{\textbf{baseline}} & \textbf{VKD} & 54.37($\pm$0.77) & 65.40($\pm$1.56) & 88.16($\pm$0.17)/87.83($\pm$0.16) & 41.46($\pm$0.89) & 86.24($\pm$0.52)/80.75($\pm$0.65)\\
    \multirow{3}{*}{} & \textbf{PKD} & 54.37($\pm$2.36) & 63.90($\pm$0.79) & 88.45($\pm$0.10)/88.08($\pm$0.06) & 41.87($\pm$1.13) & 86.37($\pm$0.60)/80.98($\pm$0.81)\\
    \multirow{3}{*}{} & \textbf{MiniLM$\dag$} & 48.45($\pm$6.19) & 61.49($\pm$0.59) & 87.82($\pm$0.08)/87.49($\pm$0.09) & 33.34($\pm$0.83) & 84.57($\pm$1.36)/78.13($\pm$2.02)\\
    \multirow{3}{*}{} & \textbf{RKD} & 51.83($\pm$5.84) & 65.22($\pm$0.90) & 88.43($\pm$0.17)/88.12($\pm$0.14) & \textbf{43.07($\pm$1.49)} & 86.87($\pm$0.47)/81.84($\pm$0.35)\\
    \hline
    \multirow{2}{*}{\textbf{proposed}}  & \multirow{2}{*}{\textbf{FSD}} & \textbf{55.49($\pm$1.61)} & \textbf{66.61($\pm$1.01)} & \textbf{88.69($\pm$0.10)}/\textbf{88.33($\pm$0.09)} & 43.03($\pm$1.43) & \textbf{87.10($\pm$0.24)}/\textbf{82.20($\pm$0.33)}\\
    & & [0.115] & [0.002] & [0.004] / [0.013] & [0.521] & [0.115] / [0.060] \\
    \hline
    \end{tabular}

    \label{table:glue}
}
\end{table*}
\footnotetext{https://huggingface.co/transformers/v2.6.0/examples.html}

\begin{table*}[ht!]
\small
{
    \centering
    \caption{Performance on large tasks of GLUE benchmark.
    Mean and standard deviation of their evaluation metrics on six runs on their development (dev.)sets.
    [\textit{p}-value]s of t-test are calculated for comparison of the proposed and the best baseline methods.
    The numbers are measured by accuracy in SST-2, QNLI, and MNLIs, and accuracy/F1 in QQP task. A bold number is defined by the same manner in Table~\ref{table:glue}.}
    \label{table:qqp&mnli}
    \begin{tabular}{clccccc}
    \hline
    & \textbf{method} & \textbf{SST-2} & \textbf{QNLI} & \textbf{QQP} & \textbf{MNLI}/\textbf{MNLI-mm} \\
    \hline
    \multirow{2}{*}{\textbf{teacher}} & \textbf{HF$^{\ast}$} & 91.97 & 87.46 & 88.40/88.31 & 90.61/81.08 \\
    \multirow{2}{*}{} & \textbf{BERT-base} & 92.09 & 91.60 & 91.07/88.06 & 84.70/84.65 \\
    \hline
    \multirow{4}{*}{\textbf{baseline}} & \textbf{VKD} & 90.92($\pm$0.62) & 88.46($\pm$0.47) & 91.03($\pm$0.08)/87.96($\pm$0.10) & 82.20($\pm$0.19)/82.85($\pm$0.12) \\
    \multirow{3}{*}{} & \textbf{PKD} & 90.77($\pm$0.41) & 88.57($\pm$0.17) & 91.00($\pm$0.11)/87.92($\pm$0.14) & 82.27($\pm$0.10)/82.57($\pm$0.21)  \\
    \multirow{3}{*}{} & \textbf{MiniLM$\dag$} & 90.23($\pm$0.39) & \textbf{89.48($\pm$0.19)} & 90.53($\pm$0.02)/87.21($\pm$0.02) & 82.23($\pm$0.09)/82.58($\pm$0.13) \\
    \multirow{3}{*}{} & \textbf{RKD} & 91.02($\pm$0.19) & 88.91($\pm$0.31) & \textbf{91.21($\pm$0.06)}/ 88.13($\pm$0.08) & 82.39($\pm$0.19)/82.92($\pm$0.13)\\
    \hline
    \multirow{2}{*}{\textbf{proposed}} & \multirow{2}{*}{\textbf{FSD}} & \textbf{91.04($\pm$0.28)} & 88.97($\pm$0.25) & 91.19($\pm$0.05)/\textbf{88.14}($\pm$0.08) & \textbf{82.42}($\pm$0.13)/\textbf{83.00($\pm$0.20)} \\
    & & [0.348] & [0.995] & [0.982] / [0.402] & [0.386] / [0.242] \\
    \hline
    \end{tabular}

}    
\end{table*}

\subsubsection{FSD method setting}
We fix the best cases of $\alpha, \tau$ and epochs on each downstream task as the result of grid search to reduce the cost of tuning hyper-parameters in FSD.
We conduct additional hyper-parameters $\beta$ set 1 in FSD (w/o $G$) and FSD methods to reduce hyper-parameter space and set $\beta$ from $\{3, 4, 5, 6\}$ except STS-B, which is set from $\{3, 4, 5, 6, 10\}$ and learning rate from $\{$3e-5, 4e-5, 5e-5, 6e-5$\}$.
Besides, by applying FSD (w/o $G$) method, we fix the best case of $\beta$ on methods for the intra-feature and local inter-feature structure methods and set $\gamma$ from $\{0.1, 0.2, \cdots, 0.9 \}$.

\subsubsection{FSD (w/o $IL$) settings}
Implementing unsupervised post-training for the global inter-feature structures, we set different memory sizes depending on a given dataset size.
In a large task such as QQP and MNLI, 300 memory entries ($|C| = 300$) are used to store centroids. 
The other smaller tasks used 100 entries ($|C| = 100$). 
The number of epochs for clustering is set by 3 for a teacher memory, which shows sufficient convergence in preliminary tests. 
After training the teacher memory, the structures in the memory are transferred to a randomly initialized student memory by distillation loss. 
Hyper-parameters for the distillation are separately set for each downstream task by greedy and grid search in terms of performance. 

First, we set $\gamma_m$ and $\beta$ from $\{$0, 1e-7, 1e-6, 1e-5, 1e-4, 1e-3,1e-2, 0.1, 1.0$\}$ find the best case, then set again $\gamma_m$ from $\{$6e-7, 7e-7, 8e-7, 9e-7, 2e-6, 3e-6, 4e-6, 5e-6$\}$ in the WNLI, RTE, and MNLI, and set from $\{$6e-5, 7e-5, 8e-5, 9e-5 2e-4, 3e-4, 4e-4, 5e-4$\}$ in the SST-2, and from $\{$6e-3, 7e-3, 8e-3, 9e-3 2e-2, 3e-2, 4e-2, 5e-2$\}$ in the QQP and set from $\{$0.1, 0.2, 0.3, 0.4, 0.5, 0.6, 0.7, 0.8, 0.9$\}$ in the STS-B, CoLA, MRPC, and QNLI.
Also, we set $\beta$ again from $\{$6e-8, 7e-8, 8e-8, 9e-8, 2e-7, 3e-7, 4e-7, 5e-7$\}$ in the STS-B, SST-2, and QQP, and set $\{$6e-7, 7e-7, 8e-7, 9e-7, 2e-6, 3e-6, 4e-6, 5e-6$\}$ in the CoLA, MRPC, QNLI, and MNLI, and set from $\{$0.1, 0.2, 0.3, 0.4, 0.5, 0.6, 0.7, 0.8, 0.9$\}$ in the WNLI, and RTE.
The $\alpha, \tau$ and epoch are fixed as the best case in VKD to reduce hyper-parameter optimization cost.

\subsubsection{FSD Loss Hyper-Parameters Settings}
We fix $\gamma_m$ and set $\gamma_g$ from $\{$1e-7, 1e-6, 1e-5, 1e-4, 1e-3,1e-2, 0.1, 1.0$\}$. 
Also, we set $\gamma_i$ from $\{$0.1, 1.0, $\beta$ of FSD (w/o $LG$)$\}$ and set $\gamma_l$ from $\{$0.1, 1.0, $\beta$ of FSD (w/o $IG$)$\}$.

\section{Results and Discussions}

\subsection{Quantitative Analysis}

\subsubsection{Model Performance}
Table~\ref{table:glue} presents the results on the GLUE dev. sets in small-sized tasks whose samples are less than 10,000.
Table~\ref{table:qqp&mnli} shows the other larger-sized tasks. 
The FSD methods show higher performance in the seven WNLI, RTE, MRPC, SST-2, STS-B, QQP, and MNLI (match and mismatch) tasks than the baseline methods, but CoLA, and QNLI show lower performance than the baseline. 
Compared to the teacher model, the proposed methods show slightly higher performance than BERT$_{BASE}$(T), by $1.0\%$ on the MRPC task.

Besides, we set null hypothesis ($\mathcal{H}_0$): $\mu_{s}= \mu_{\texttt{FSD}}$, where $\mu_s$ is a mean of the best case of baseline samples, and $\mu_{\texttt{FSD}}$ is a mean of FSD results.
We estimate right-tail \textit{p}-value if $\mu_{s} \leq \mu_{FSD}$, else left-tail \textit{p}-value.
By the \textit{p}-value results, FSD shows statistically significant on RTE, STS-B, and MRPC. 
In the results, the FSD method effectively and stably improves the test accuracy of the benchmark, especially on small datasets. 
Furthermore, the methods can generalize the student model to show better performance than the teacher models in particular tasks (MRPC and QQP).

\begin{figure*}[t!]
\centering
    \begin{subfloat}{\includegraphics[width=0.24\textwidth, height=2.8cm]{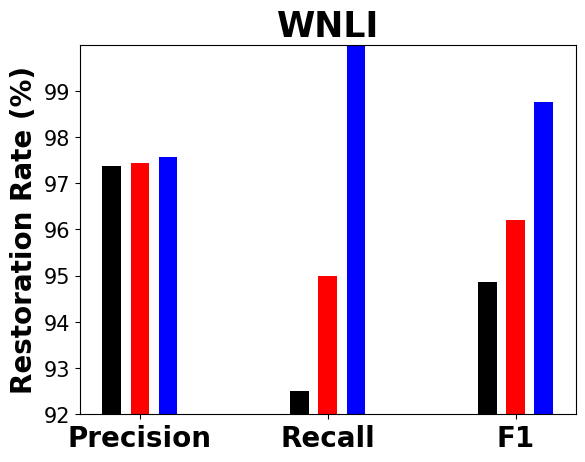}}
    \end{subfloat}%
    \begin{subfloat}{\includegraphics[width=0.24\textwidth, height=2.8cm]{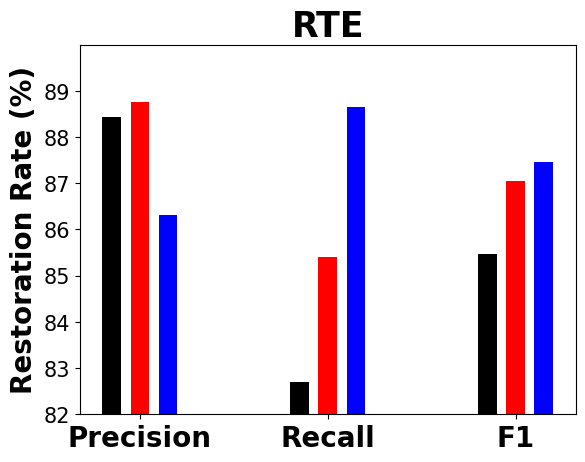}}
    \end{subfloat}%
    \begin{subfloat}{\includegraphics[width=0.24\textwidth, height=2.8cm]{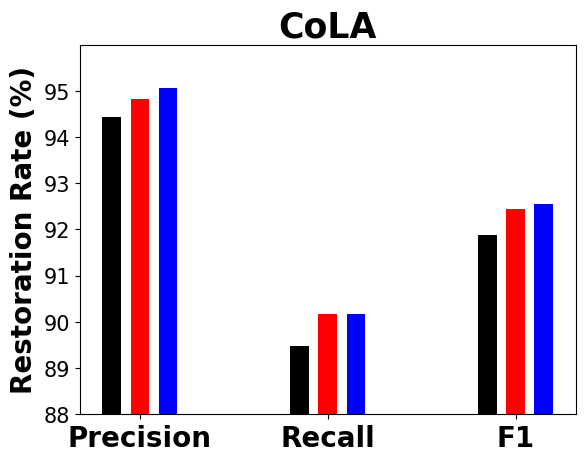}}
    \end{subfloat}%
    \begin{subfloat}{\includegraphics[width=0.24\textwidth, height=2.8cm]{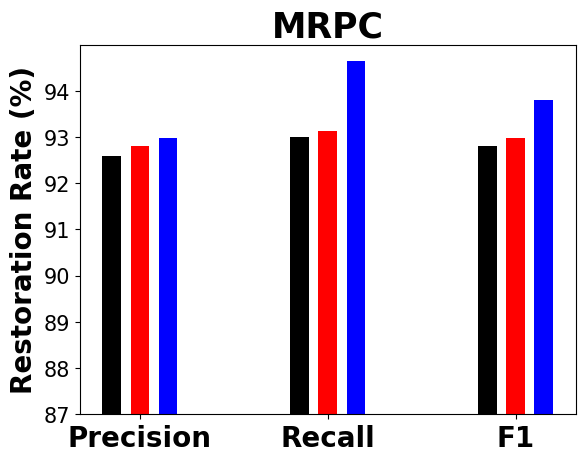}}
    \end{subfloat}%
    \hfill
    \begin{subfloat}{\includegraphics[width=0.24\textwidth, height=2.8cm]{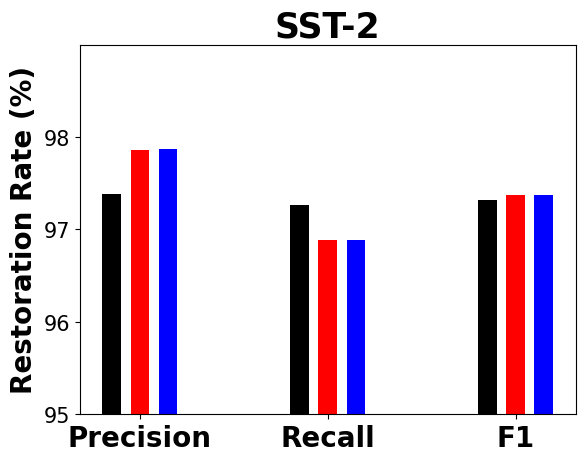}}
    \end{subfloat}%
    \begin{subfloat}{\includegraphics[width=0.24\textwidth, height=2.8cm]{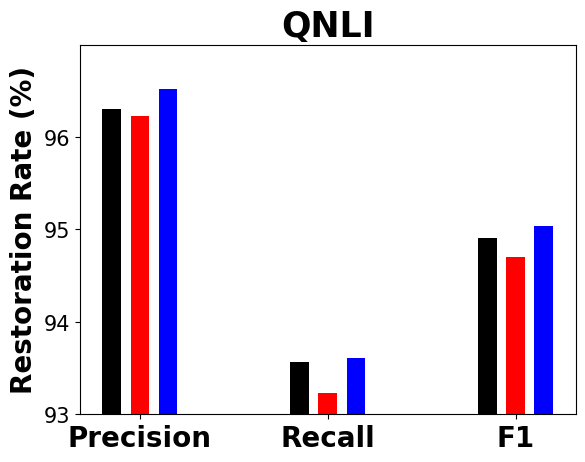}}
    \end{subfloat}%
    \begin{subfloat}{\includegraphics[width=0.24\textwidth, height=2.8cm]{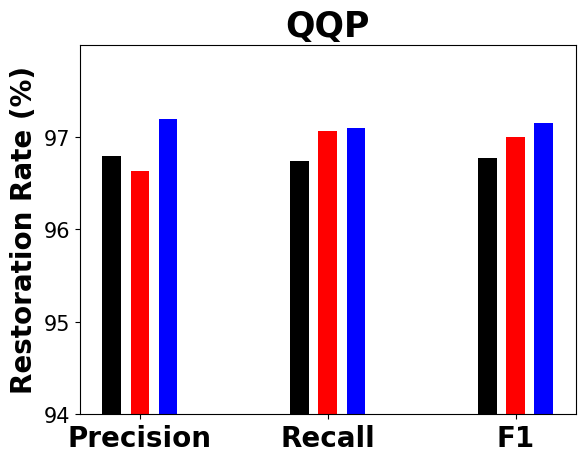}}
    \end{subfloat}%
    \begin{subfloat}{\includegraphics[width=0.24\textwidth, height=2.8cm]{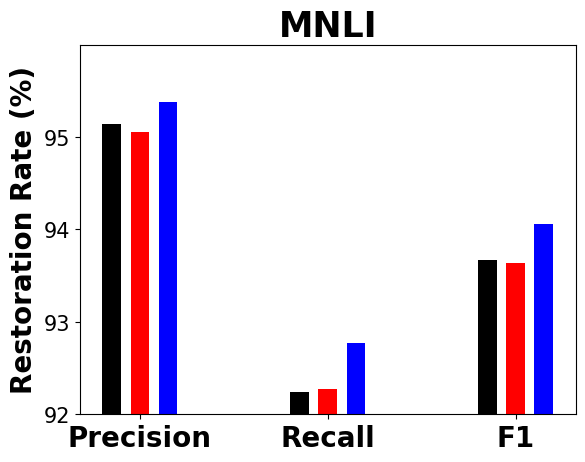}}
    \end{subfloat}%
    \hfill
    \begin{subfloat}{\includegraphics[width=0.95\textwidth]{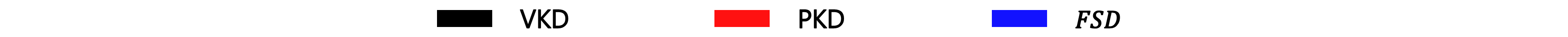}}
    \end{subfloat}
    \caption{Restoration rate of teachers' predictions in students on the GLUE tasks. MNLI used the matched subset.}
    \label{fig:restoration score}
\end{figure*}

\subsubsection{Ablation Study}

\begin{table*}[h!]
\tiny
{
    \centering
    \caption{
    Ablation study on GLUE benchmark. The performance reduction rates over FSD 
    ($\frac{\textrm{reduced performance}}{\textrm{FSD performance}}$) are shown.
    The results are averaged over six runs with different random seeds.
    \textbf{Avg.} is a average of results.
    The numbers follow same manner on Table~\ref{table:glue} and~\ref{table:qqp&mnli}.}
\label{table:glue_abla}
    \begin{tabular}{clccccc}
    \hline
    & \textbf{method} & \textbf{FSD} (w/o $LG$) & \textbf{FSD} (w/o $IG$) & \textbf{FSD} (w/o $IL$) & \textbf{FSD} (w/o $G$) \\
    \hline
    
    \multirow{9}{*}{\textbf{tasks}} & \textbf{WNLI} &  -6.60$\%$ & -1.02$\%$ & -3.55$\%$ & -2.03$\%$\\
    \multirow{9}{*}{} & \textbf{RTE} &  -0.72$\%$ & -0.45 $\%$ & -0.18$\%$ & -0.63$\%$ \\
    \multirow{9}{*}{} & \textbf{CoLA} &  -2.15$\%$ & -8.14$\%$ & -1.60$\%$ & -2.50$\%$\\
    \multirow{9}{*}{} & \textbf{SST-2} &  -0.44$\%$ & -0.57$\%$ & -0.31$\%$ & -0.34$\%$\\
    \multirow{9}{*}{} & \textbf{QNLI} &  -0.14$\%$ & -0.56$\%$ & -0.51$\%$ & -0.31$\%$\\
    \multirow{9}{*}{} & \textbf{STS-B} &  -0.04$\%$ / -0.02$\%$ & -0.03$\%$ / -0.08$\%$ & -0.21$\%$ / -0.22$\%$ & -0.02$\%$ / -0.03$\%$\\
    \multirow{9}{*}{} & \textbf{MRPC} &  -0.18$\%$ / -0.40$\%$ & -0.10$\%$ / -0.24$\%$ & -1.24$\%$ / -2.16$\%$ & -0.08$\%$ / -0.24$\%$\\
    \multirow{9}{*}{} & \textbf{QQP} & -0.03$\%$ / -0.06$\%$ & -0.00$\%$ / +0.01$\%$ & -0.20$\%$ / -0.23$\%$ & -0.07$\%$ / -0.10$\%$ \\
    \multirow{9}{*}{} & \textbf{MNLI} & -0.09$\%$ / -0.18$\%$ & -0.08$\%$ / -0.47$\%$ & -0.46$\%$ / -0.27$\%$ & +0.36$\%$ / -0.36$\%$\\
    \hline
    & \textbf{Avg.} & -0.78$\%$ & -0.90$\%$ & -0.86$\%$ & -0.48$\%$ \\
    \hline
    \end{tabular}

}
\end{table*}
As shown in Table~\ref{table:glue_abla}, considering all structure results are higher than other methods.
When we remove single structure method (w/o LG, w/o IG, and w/o IL), FSD (w/o LG) shows less decline and FSD (w/o IG) shows worst case among three cases.
In addition, FSD (w/o G) case shows better than single structure distillation methods.
These results imply that integrated methods are better than single structure distillation, and in particular, applied for all feature structure transfers teacher's knowledge is more effective than others.
In particular, the gap between FSD and other methods on small datasets such WNLI, and CoLA is larger than other tasks.
These results show that the proposed method is more effective on small datasets.

\begin{table}[!h]
    \centering
    \caption{Each value represents the standard deviation of model performance of 18 runs composed 6 runs for three mini-batch sizes in $\{$8, 16, 32$\}$.} 
    \label{tab:mini-batch stdev}
    \begin{tabular}{l|c|c}
        \hline
         \textbf{method} & \textbf{MRPC} & \textbf{SST-2} \\
         \hline
         \textbf{FSD} (w/o $IG$) & 0.66/0.44 & 0.68 \\
         \textbf{FSD} (w/o $IL$) & \textbf{0.47/0.38} & \textbf{0.62} \\
         \hline
    \end{tabular}
\end{table}
\subsubsection{Stability of Global Structure for Transferring}
We estimate of standard deviation in different mini-batch cases to show that utilized memory method (FSD (w/o $IL$)) is less affected by mini-batch size.
We select MRPC and SST-2 tasks because each task is in small and large dataset of the GLUE and both tasks are stably improved model performance.
As shown in Table~\ref{tab:mini-batch stdev}, two tasks standard deviation is the lowest in FSD (w/o $IL$).
These results imply that regardless of mini-batch size, FSD (w/o $IL$) method preserves teacher feature structure information in memory and stably maintains model performance.

\subsection{Qualitative Analysis}

\subsubsection{Restoration Rate of Teacher Prediction}



Fig.~\ref{fig:restoration score} shows the restoration rates of teacher prediction. 
As all tasks are binary classification except STS-B, we plot precision, recall, and F1-score for each method in each task, over teacher prediction result as the ground-truth.
FSD method generally shows higher restoration rates than baseline.

In KD, correctly recovering a teacher's prediction affects the quality of transferring.
In the restoration rate results analysis, the higher restoration rate implies student imitates teacher results accurately, therefore the proposed method is effective to emulate teachers. 
Moreover, FSD shows better mimicry of teacher results than the other proposed methods.
As shown in the WNLI, RTE, CoLA, and MRPC results, FSD method significantly shows more effective on small datasets than larger datasets.


\begin{figure*}
    \centering
    \begin{subfloat}[]{\includegraphics[width=0.49\textwidth]{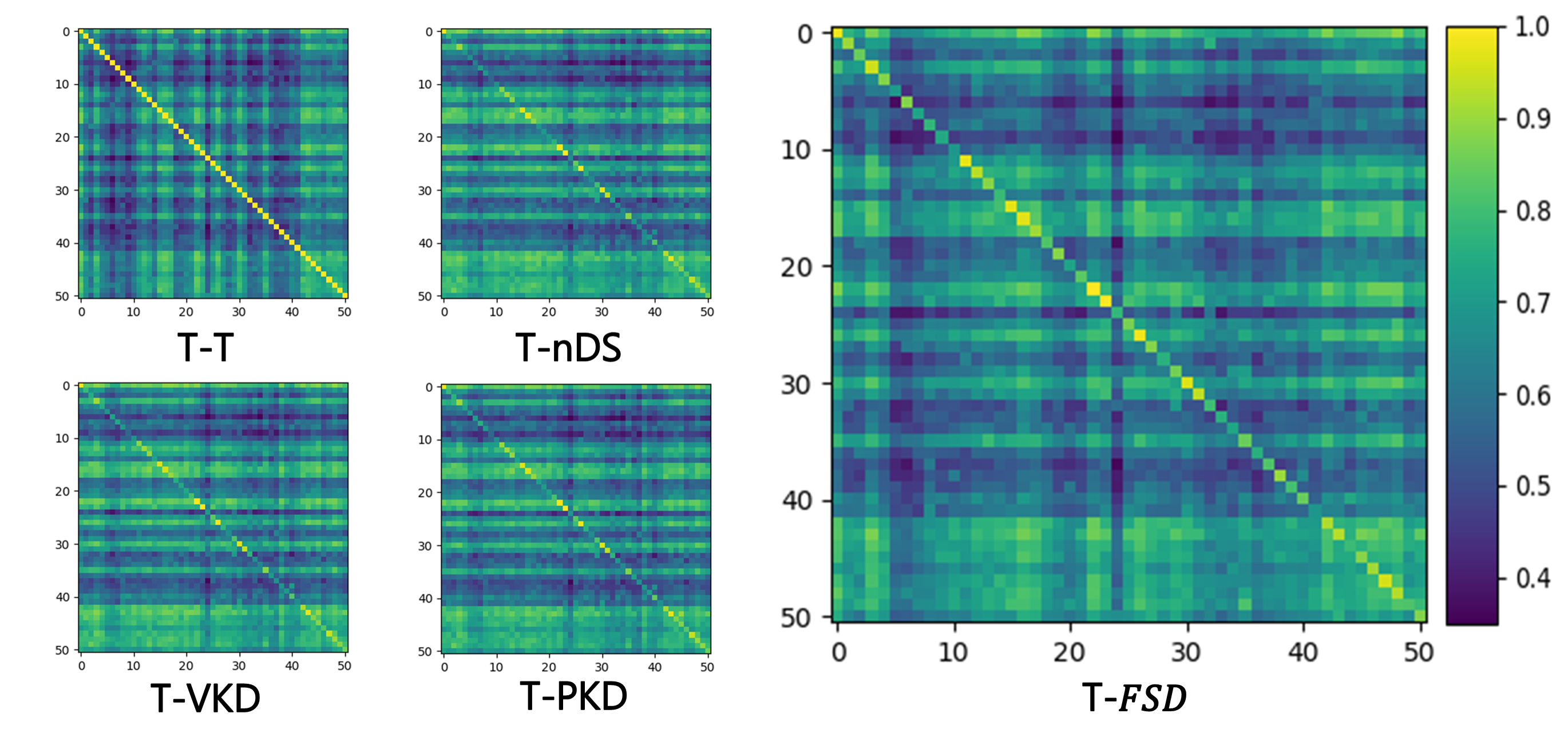}}
    \end{subfloat}
    \hfill
    \begin{subfloat}[]{\includegraphics[width=0.49\textwidth]{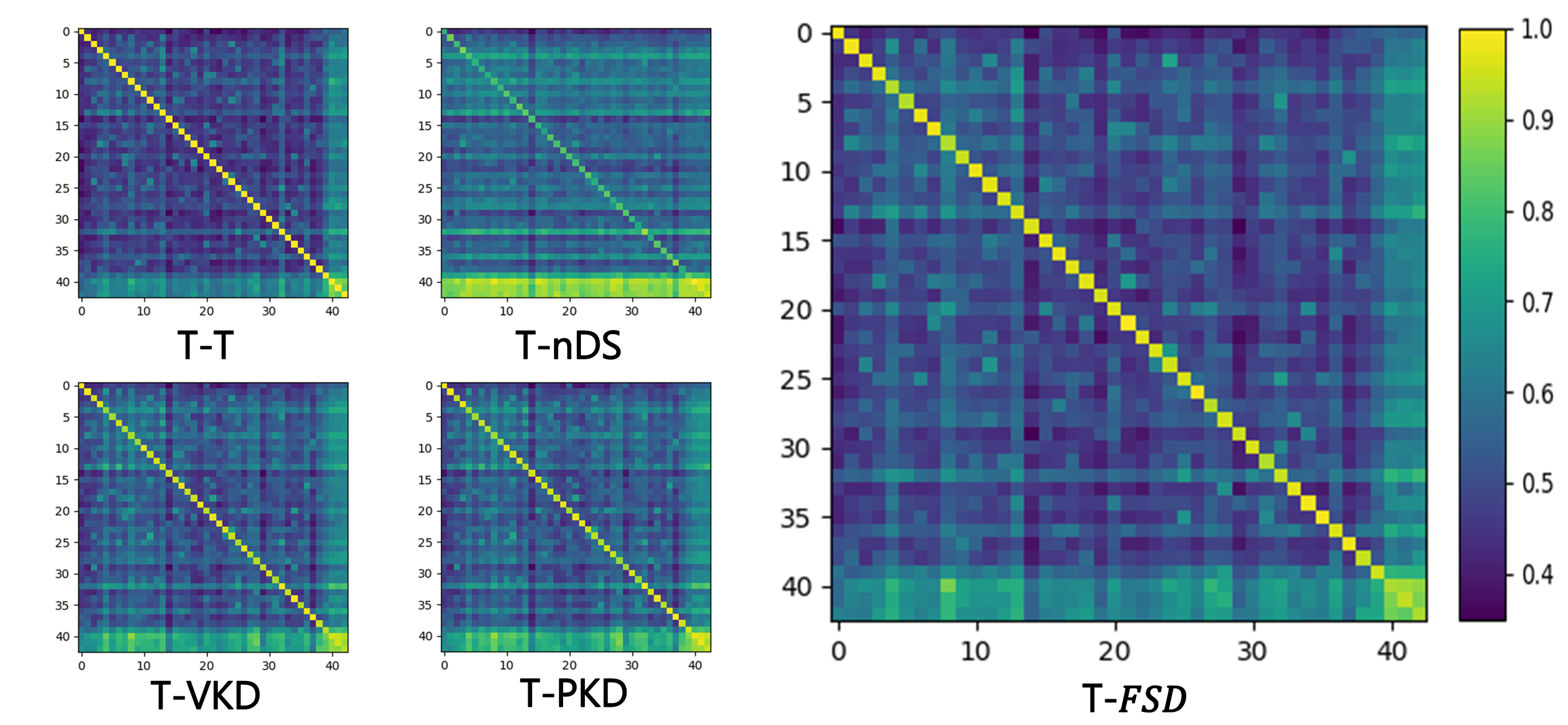}}
    \end{subfloat}
    \caption{CKA heat maps of GLUE benchmark. CoLA data set result is representative on GLUE benchmark, x-axis, y-axis are indices of mini-batches on the same data set. Each pixel shows the CKA similarity between teacher's and other model's representations generated from their mini-batches. In the diagonal lines, the same samples are used for the generation. Right side bar shows the normalized range of CKA similarities observed in each task. (a) is a CoLA case and (b) is a SST-2 case. The heat maps of all the other tasks are in Appendix.}
    \label{fig:cka_heatmap}
\end{figure*}

\begin{table*}[t]
\centering
\caption{Each value is the average of diagonal values for each task and each teacher-student pair. 
}
\label{table:heatmap}
\begin{tabular}{clccccccccc|c}

\multicolumn{12}{c}{\small CKA heat map diagonals}\\
\hline
 & \textbf{method}& \textbf{WNLI} & \textbf{RTE} & \textbf{STS-B} & \textbf{CoLA} & \textbf{MRPC} & \textbf{SST-2} & \textbf{QNLI} & \textbf{QQP} & \textbf{MNLI} & \textbf{Avg} \\
 \hline
 \multirow{2}{*}{\textbf{No KD}} & \textbf{T-T} & 1.000 & 1.000 & 1.000 & 1.000 & 1.000 & 1.000 & 1.000 & 1.000 & 1.000 & 1.000\\
 \multirow{2}{*}{} & \textbf{T-nDS} & 0.998 & 0.985 & 0.954 & 0.803 & 0.981 & 0.794 & 0.972 & 0.909 & 0.947 & 0.927\\
 \hline
 
 \multirow{2}{*}{\textbf{baseline}} & \textbf{T-VKD} & 0.997 & 0.984 & 0.955 & 0.811 & 0.979 & 0.910 & 0.965 & 0.889 & 0.919 & 0.934\\
 \multirow{2}{*}{} & \textbf{T-PKD} & 0.997 & 0.980 & 0.956 & 0.813 & 0.979 & 0.900 & 0.961 & 0.903 & 0.944 & 0.937\\
 \hline
 \textbf{proposed} & \textbf{T-}$\textbf{FSD}$ & 0.999 & 0.990 & 0.981 & 0.835 & 0.990 & 0.942 & 0.985 & 0.958 & 0.942 & \textbf{0.958} \\
 \hline
\end{tabular}
\end{table*}

\begin{figure}
    \centering
    \includegraphics[width=0.47\textwidth]{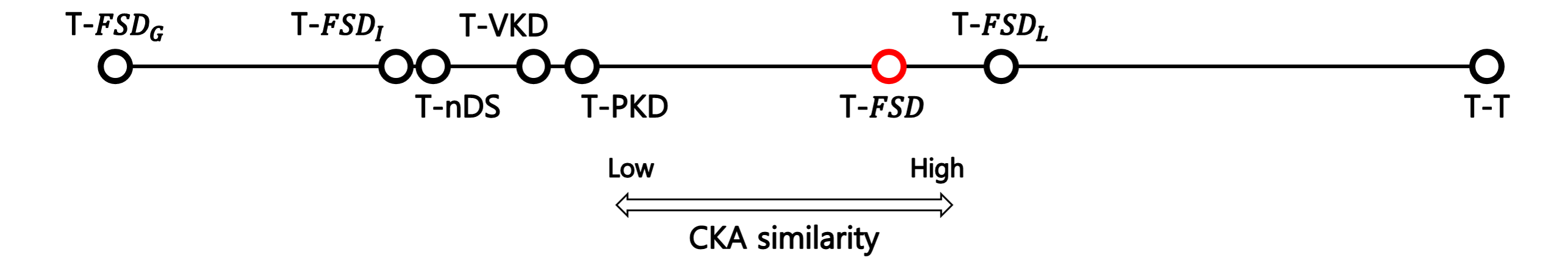}
    \caption{Each point represents the average of diagonal values on CKA heat map. The average of T-FSD$_G$ is 0.905 and T-T case is 1.000. We plot each point to reflect difference.}
    \label{fig:lines}
\end{figure}

\subsubsection{Task-Specific Structural Similarity Between Teacher and Student}
\label{subsec:structural similariy}
We evaluate CKA similarity heat maps of teacher by teacher (T-T), teacher by no-Distillation-Student (T-noDS), and teacher by student with methods for each task to investigate structural properties on CKA perspective.
Fig.~\ref{fig:cka_heatmap} shows CKA heat maps and Table~\ref{table:heatmap} shows the average of similarities on the diagonal lines and the average over all tasks.
CKA similarity evaluation is conducted on mini-batches. 
We split a test set to build a mini-batch pool for each task. Then, teacher and another model separately select their mini-batch and generates corresponding feature for evaluating CKA similarity. This value is shown in a pixel and we repeated it for all mini-batch pairs. The size of mini-batches differs by tasks because of different test set size. 

Fig.~\ref{fig:cka_heatmap}, the lightness of each diagonal line implies  the accuracy of transferring teacher knowledge captured by CKA. Its maximum value is 1.0 obtained in any T-T cases.
More accurate numerical comparison in Table~\ref{table:heatmap} shows that  FSD shows significantly better average diagonal values over all tasks than other methods.

In the comparison of heat map patterns, the FSD is similar to the T-T heat map, called teacher group. 
In contrast, VKD, and PKD are more closer to noDS, called noDS group.
In the results, the different range of heat maps and patterns show the implicit difference of structures between tasks. 
Depending on the complexity of transferring the teacher's structures and conflict with students' structures, the performance is heavily affected.
The clear distinction of heat map patterns between the teacher and noDS groups implies that the transferred structures largely differ by their types.

    
    

\subsubsection{Impact of CKA to model performance}
As shown in Fig.~\ref{fig:lines}, VKD, PKD, FSD$_I$, and FSD$_G$ are close to nDS, and FSD$_L$ and FSD are located more closer to ideal case (T-T).
It shows that VKD and PKD less reflect teacher's knowledge, compare to FSD$_L$ and FSD.
In addition, FSD is interpolated by FSD$_I$, FSD$_L$, and FSD$_G$.

\begin{figure*}
    \centering
    \includegraphics[width=0.99\textwidth]{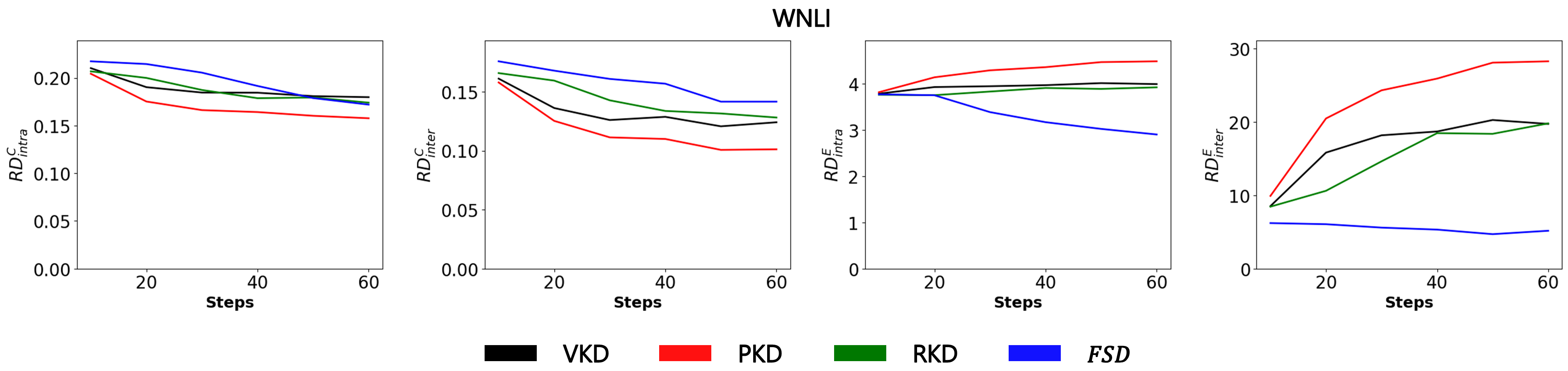}
    \caption{The results of relation difference. Each task shows the results of using different four difference metrics: $RD^C_{intra}$, second is $RD^C_{inter}$, third is $RD^E_{intra}$, and $RD^E_{inter}$. Fisrt row, WNLI, results are representative. The results of all the other tasks are in Appendix.}
    \label{fig:RD_results}
\end{figure*}

As shown in table~\ref{table:rank}, the average of FSD $RD$ rank is the highest, but $RD$ structure graph pattern shown in Fig.~\ref{fig:RD_results} is not consistent in all GLUE task.
On the other hand, CKA heat map diagonals in Fig.~\ref{fig:cka_heatmap} are consistent regardless of tasks.
As referred on the Results section, CKA heat map could be a clue to explain the best case of method (FSD).
Therefore, CKA analysis is more related metric than Euclidean distance and cosine similarity to explain KD model performance.

\begin{table*}[htbp]
\centering
\caption{Each value is the average of ranks of four \textit{relation difference} results of each task. The lower is the better and the final difference values after training are used for the ranking.  (Avg: the average of the average ranks over all tasks).}
\label{table:rank}
\begin{tabular}{clccccccccc|c}
\multicolumn{12}{c}{\bf{Average Rank of Relation Difference Table}}\\
\hline
 & \textbf{method}& \textbf{WNLI} & \textbf{RTE} & \textbf{STS-B} & \textbf{CoLA} & \textbf{MRPC} & \textbf{SST-2} & \textbf{QNLI} & \textbf{QQP} & \textbf{MNLI} & \textbf{Avg} \\
 \hline
 \multirow{3}{*}{\textbf{baseline}} & \textbf{VKD} & 3.00 & 2.50 & 3.00 & 3.00 & 3.25 & 3.50 & 2.75 & \textbf{2.50} & 2.50 & 2.89\\
 \multirow{3}{*}{} & \textbf{PKD} & 2.50 & 2.50 & 2.50 & 2.75 & 2.50 & 3.25 & 2.75 & \textbf{2.50} & 2.75 & 2.64 \\
 \multirow{3}{*}{} & \textbf{RKD} & 2.50 & 3.00 & 3.00 & 3.00 & 2.50 & \textbf{1.50} & \textbf{2.00} & \textbf{2.50} & \textbf{2.25} & 2.44 \\
 \hline
 \textbf{proposed} & \textbf{FSD} & \textbf{2.00} & \textbf{2.00} & \textbf{1.50} & \textbf{1.25} & \textbf{2.25} & 1.75 & 2.50 & \textbf{2.50} & 2.50 & \textbf{2.03}\\
 \hline
\end{tabular}
\end{table*}

\subsection{Additional Discussion}
In this subsection, we discuss about additional observation from the above analysis results.

\subsubsection{Patterns of Transferring Structures}

To analyze the patterns of transferred structures, we evaluate \textit{relation difference} ($RD$) for inter-feature structure ($RD_{inter}$) as
\begin{equation}
RD_{inter}=\sum_{i,j} \frac{|\psi(\displaystyle \mH^T_{i}, \displaystyle \mH^T_{j})-\psi(\displaystyle \mH^S_{i}, \displaystyle \mH^S_{j})|}{|B|^2},\\
 \label{eq: relation difference inter}
\end{equation}
where $i,j \in \{1,2,\cdots,|B|\}$,
and for intra-feature structure ($RD_{intra}$) as
\begin{equation}
RD_{intra}=\sum_{i,j,k}
 \frac{ |\psi(\displaystyle \mH^T_{i,j}, \displaystyle \mH^T_{i,k})-\psi(\displaystyle \mH^S_{i,j}, \displaystyle \mH^S_{i,k})|}{|W|^2\cdot|B|},\\
 \label{eq: relation difference intra}
\end{equation}
where $i \in \{$1,2, $\cdots$, $|B|\}$ and $ j,k \in \{$1,2, $\cdots$, $|W|\}$.
This metric implies the average difference of the relation unit building intra-feature structures and inter-feature structures between the teacher and students.
$RD^E$ and $RD^C$ are defined by the same manner of $\textbf{F}^E_{\mathbf{Mh}}$ and $\textbf{F}^C_{\mathbf{Mh}}$ to specify $\psi$.
Then, we evaluate the rank of the last iteration $RD$ values over each method on each task.
Baseline models and primitive FSD methods are tested for clear analysis of the impact of structure types.

$RD$ values in training are illustrated in Fig.~\ref{fig:RD_results}, and their $RD$ ranks are shown on the Table~\ref{table:rank}. The lower $RD$, the better rank close to one.
In the WNLI graph, most proposed methods were more effective to reduce $RD^E$ than the baselines, while  $RD^C$ shows inconsistent superiority.
The pattern is similarly observed in the other GLUE tasks.
In Table~\ref{table:rank}, the proposed FSD method shows the best ranks on most tasks.
FSD shows the best average rank on the GLUE task, where FSD is the first rank in the WNLI, RTE, STS-B, CoLA, MRPC, and QQP and the second-best rank in the rest of GLUE tasks.



The $RD$ results of WNLI task show that the proposed method preserves structures on $RD^E$ but unstably transfers the structures on $RD^C$ because of the CKA property to preserve Euclidean distance and dot product.

The higher ranks of FSD method in most tasks implies that CKA similarity is effective to preserve structures on $RD$ metrics.
RKD, the best among baseline, is still significantly worse than FSD even if it spends larger computational cost for evaluating $O(n^3)$ relations than $O(n^2)$ of FSD method. 
In sum, FSD using CKA similarity effectively transfers teachers' potential structures defined by various difference metrics with relatively good computational efficiency.

\begin{table}[]
    \centering
    \caption{The number of model parameters ($\#$Param), and training and inference time ratio compare to VKD, \textit{i.e.}) $t_{\text{methods}} / t_{\text{VKD}}$, where $t$ is the training or inference time.}
    \label{tab: model and time complexity}
    \begin{tabular}{c|c|c|c}
    \hline
        Method & $\#$Param & Training $\uparrow$ & Inference $\uparrow$ \\
        \hline
        VKD & 6.70M & 1.00 $\times$ & 1.00 $\times$ \\
        PKD & 6.70M & 0.96 $\times$ & 1.00 $\times$ \\
        RKD & 6.70M & 0.77 $\times$ & 1.00 $\times$ \\
        \hline
        FSD & 7.68M & 0.92 $\times$ & 0.88 $\times$ \\
        \hline
    \end{tabular}
\end{table}

\subsubsection{Model and Time Complexity}
As shown in Table~\ref{tab: model and time complexity} we conduct model and time complexity for training and inference.
Even though our proposed method model size is about 1.15 times larger than RKD, the FSD is much faster than RKD during training because the observed structure size is $O(n^2)$ and $O(n^3)$ in FSD and RKD, respectively.

\section{Conclusion}
In this paper, we addressed transferring rich information of feature representations via knowledge distillation in BERT. 
To represent the features, we proposed three levels of feature structures defined by CKA similarity: intra-feature, local inter-feature, and global inter-feature structures. 
To transfer them, we implemented \textbf{feature structure distillation} methods separately for the structures, especially for the global structures using memory-augmented transferring with clustering. 
We could find that transferring all the structures induces more similar student representations to its teacher and consistently improves performance in the GLUE language understanding tasks.
This work can be extended to more downstream applications using pre-trained BERT for more fine transferring.

\section*{Acknowledgment}
This work was supported by the National Research Foundation of Korea (NRF) grant funded by the Korea government (MSIT) (2022R1A2C2012054).











\bibliographystyle{cas-model2-names}

\bibliography{cas-refs}




\clearpage

\section*{Appendix}
\subsection{CKA Heat Maps of All GLUE Benchmark Results}
This material is to show the full results of Fig.~\ref{fig:cka_heatmap} in the manuscript. 

\begin{figure*}[b!]
    \centering
    \includegraphics[width=1.0 \textwidth]{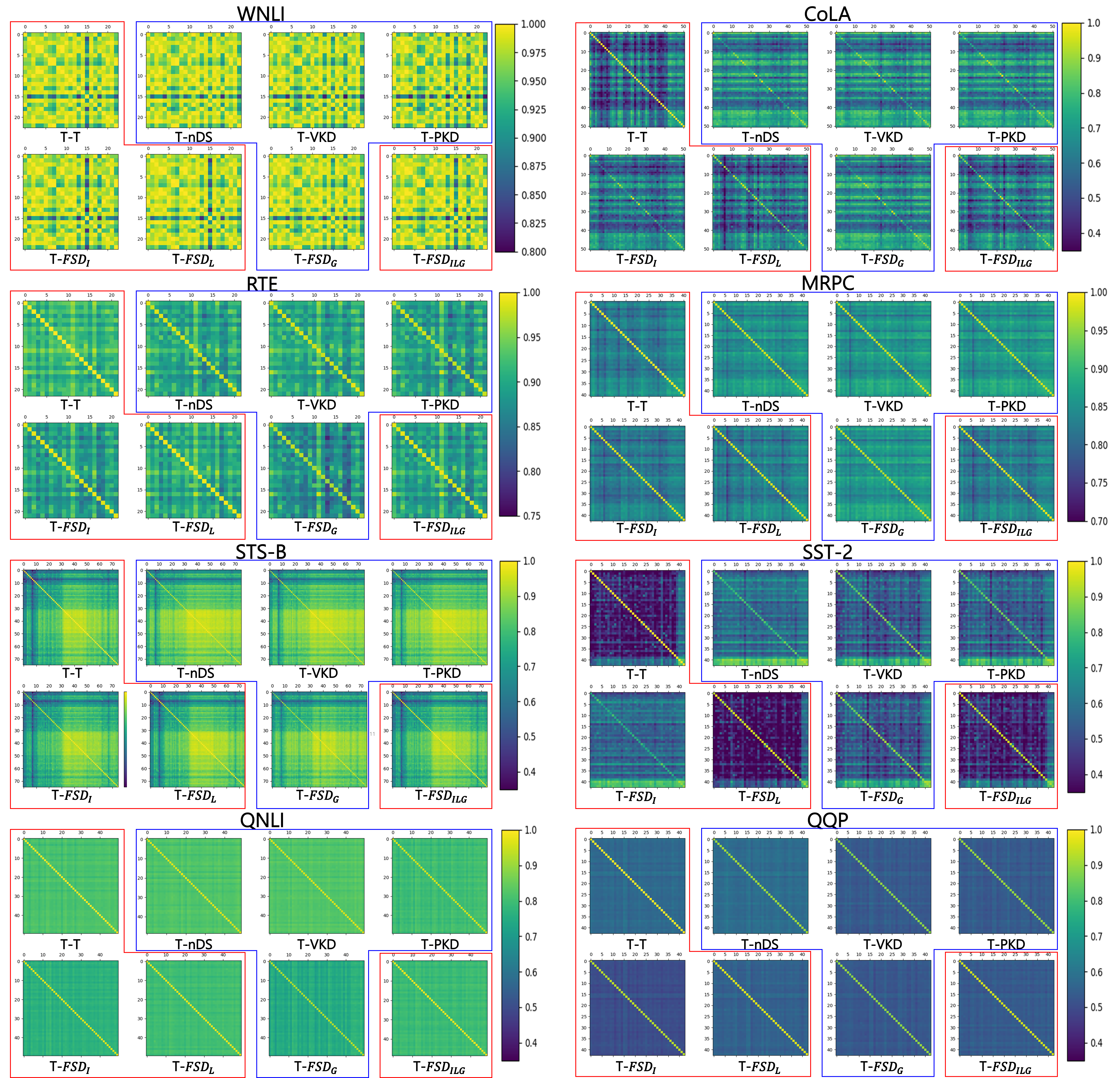}
    \caption{CKA heatmap of GLUE benchmark. Red box represents teacher closed methods and blue represents no-Distillation-Student closed methods.}
    \label{fig:cka heat map}
\end{figure*}

As shown in Fig.~\ref{fig:cka heat map}, overall diagonal lines are lighter in WNLI, RTE, STS-B, MRPC, QNLI, and QQP than CoLA, SST-2, and MNLI, which shows the task-specific difference between teacher and student knowledge.
Normally, patterns are divided into teacher closed (FSD$_I$, FSD$_L$, and FSD) and no-Distillation-Student closed models (VKD, PKD, and FSD$_G$), and it is clearly shown on CoLA, RTE, MRPC, and SST-2.
These results still show that the proposed method is more effective on small datasets.

\subsection{Relation Difference of All GLUE Benchmark Result}
\begin{figure*}[ht]
    \centering
    \includegraphics[width=0.99\textwidth]{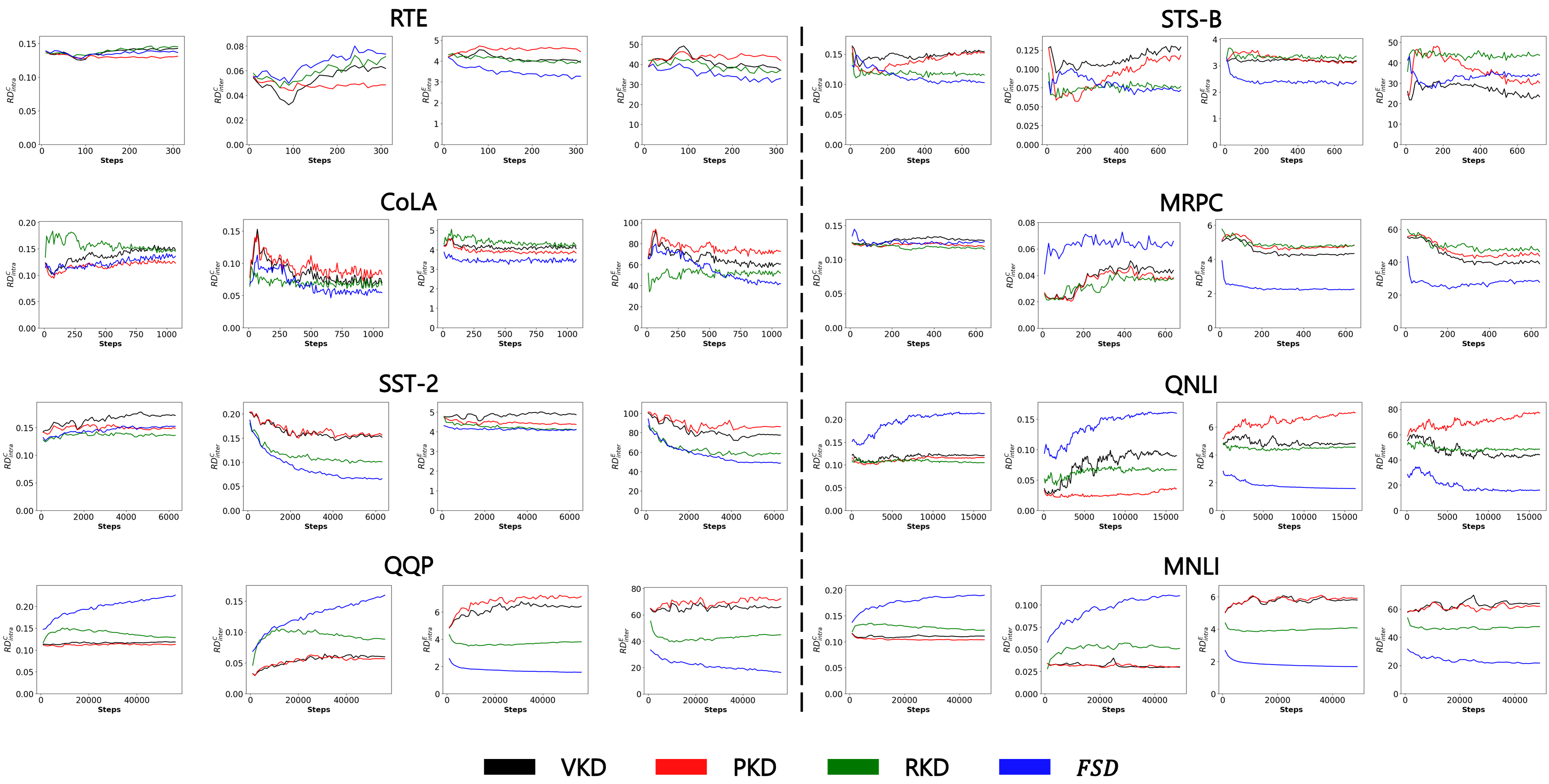}
    \caption{The results of relation difference. Each task shows the results of using different four difference metrics: $RD^C_{intra}$, second is $RD^C_{inter}$, third is $RD^E_{intra}$, and $RD^E_{inter}$.}
    \label{fig:RD_results_appendix}
\end{figure*}

This material is to show the full results of Fig.~\ref{fig:RD_results} in the manuscript. 
In Fig.~\ref{fig:RD_results_appendix}, similar patterns are shown in overall tasks. 
In some cases as COLA, SST-2, and STS-B, Euclidean distance also largely decreases in FSD compared to baselines. 

The distinctive patterns in FSD implies that the knowledge transferred by it is all different, which explains why the integration shows the best performance.

\end{document}